\pdfoutput=1

\documentclass[11pt]{article}

\usepackage{styles/acl}

\usepackage{times}
\usepackage{latexsym}
\usepackage[T1]{fontenc}
\usepackage[utf8]{inputenc}
\usepackage{microtype}
\usepackage{inconsolata}
\usepackage{graphicx}

\usepackage{caption}
\usepackage{subcaption}

\newcommand{\reffig}[1]{Figure~\ref{#1}}
\newcommand{\reftbl}[1]{Table~\ref{#1}}
\newcommand{\refsec}[1]{Section~\ref{#1}}

\title{Reducing Hallucinations in Language Model-based SPARQL Query Generation Using Post-Generation Memory Retrieval}

\author{
 \textbf{Aditya Sharma\textsuperscript{1,2}},
 \textbf{Christopher J. Pal\textsuperscript{1,2,3},
 \textbf{Amal Zouaq\textsuperscript{1,2}}}
\\
\\
 \textsuperscript{1}Polytechnique Montréal,
 \textsuperscript{2}Mila - Quebec AI Institute,
 \textsuperscript{3}Canada CIFAR AI Chair
\\
 \small{
   \textbf{Correspondence:} \href{mailto:aditya.sharma@mila.quebec}{aditya.sharma@mila.quebec}
 }
}

\begin{document}
\maketitle

\begin{abstract}

The ability to generate SPARQL queries from natural language questions is crucial for ensuring efficient and accurate retrieval of structured data from knowledge graphs (KG). While large language models (LLMs) have been widely adopted for SPARQL query generation, they are often susceptible to hallucinations and out-of-distribution errors when generating KG elements, such as Uniform Resource Identifiers (URIs), based on opaque internal parametric knowledge. 
We propose PGMR (Post-Generation Memory Retrieval), a modular framework where the LLM produces an intermediate query using natural language placeholders for URIs, and a non-parametric memory module is subsequently employed to retrieve and resolve the correct KG URIs. PGMR significantly enhances query correctness (SQM) across various LLMs, datasets, and distribution shifts, while achieving the near-complete suppression of URI hallucinations. 
Critically, we demonstrate PGMR's superior safety and robustness: a retrieval confidence threshold enables PGMR to effectively refuse to answer queries that lack support, and the retriever proves highly resilient to memory noise, maintaining strong performance even when the non-parametric memory size is scaled up to 9 times with irrelevant, distracting entities.

\end{abstract}
\section{Introduction}
\label{sec:introduction}

Question answering (QA) is a fundamental task in natural language processing research. Many QA systems leverage knowledge graphs (KGs) \citep{saxena2020improving, kgqa_survey, kg_survey}, which represent world knowledge as multi-relational graphs of factual triples \textit{(subject, relation, object)}, for example, \textit{(The Truman Show, nominated for, Academy Award for Best Supporting Actor)}.
Answering questions over KGs (KGQA) can typically involve two steps: first, translating a natural language question into its corresponding SPARQL\footnote{\url{https://www.w3.org/TR/rdf-sparql-query/}} query (discussed in \refsec{sec:sparql_queries}), and second, executing that query against the KG to retrieve an answer \citep{banerjee}.

Large language models (LLMs) have shown promise for this task \citep{meyer2024assessing, karou2023, banerjee, qi2024enhancing, Text2SPARQL}.
However, the process of authoring SPARQL queries remains demanding for non-experts and is non-trivial even for modern LLMs, as it mandates an intimate familiarity with the underlying KG schema and the obscure structure of its Uniform Resource Identifiers (URIs - described in \refsec{sec:uri}). Specifically, the use of non-semantic identifiers within platforms such as Wikidata (e.g., a prefix followed by arbitrary digits like Q937 for Albert Einstein) forces LLMs to rely solely on their parametric memory for accurate identifier mapping, impeding robust query generation.
Furthermore, LLMs often hallucinate and produce plausible-looking but incorrect identifiers, such as URIs that don't exist in the KG \citep{hallucination_survey_1, hallucination_survey_2, huang2021factual}.
This limitation raises concerns about the reliability of LLMs in real-world information retrieval (IR) applications and has driven extensive research into techniques for detecting and mitigating such errors \citep{lin2024towards, varshney2023stitch, dhuliawala2023chain, chern2023factool, li2023halueval}.

Retrieval-augmented generation (RAG) \citep{knnlm, rag_og, guu2020retrieval, lewis2020retrieval} is a popular paradigm that is often employed for reducing hallucinations \citep{hallucination_survey_1}. It uses a non-parametric memory for retrieving information in response to a question. This information is provided in the prompt for the LLM before generation to enhance the grounding of responses.
However, since the LLM is not strictly constrained to rely on the retrieved context, it may still generate responses that are not fully grounded, thereby failing to eliminate hallucinations \citep{barnett2024seven}.

In response to these limitations, we propose PGMR (Post-Generation Memory Retrieval) \footnote{Code and data available at \url{https://github.com/
Lama-West/PGMR}}, a new modular architecture featuring non-parametric memory modules with a retriever for managing KG elements. This approach enables LLMs to focus on generating SPARQL query syntax while the retriever ensures accurate retrieval of relevant KG elements.
PGMR enables the LLM to generate intermediate queries with a structured query representation that preserves SPARQL syntax while replacing KG URIs with natural language placeholders.
Each placeholder is paired with the corresponding URI label and description in a mapping block.
This design enables LLMs to focus on generating syntactically correct query structures, while deferring the grounding of placeholders to precise URIs to a post-generation retrieval step.

The core conceptual innovation of PGMR lies in its temporal repositioning of the retrieval phase. While RAG relies on pre-generation retrieval from natural language questions that often lack semantic alignment with URI metadata in the memory, PGMR performs retrieval after generation and leverages the LLM to generate explicit labels and descriptions after the query structure is formed. By producing these placeholders in a format that directly mirrors the non-parametric memory, PGMR achieves significantly higher latent similarity compared to RAG. This architecture effectively decouples structural synthesis from identifier grounding, ensuring that all generated URIs are strictly aligned with existing knowledge graph elements to nearly eliminate hallucinations.

We address the following research questions:
\begin{enumerate}
    \item Does separating query structural prediction from identifier grounding reduce URI hallucinations while improving query correctness? 
    \item How effectively does a retrieval confidence threshold enable accurate query refusal under incomplete knowledge conditions?
    \item How robust is PGMR's performance when subjected to substantial memory scale-up and irrelevant data injection?
\end{enumerate}

This study introduces several key contributions:
\begin{enumerate}
    \item We propose PGMR, a modular architecture that enhances LLM-based SPARQL generation by explicitly separating query syntax (generated by the LLM) from URI resolution (handled by a non-parametric retriever).
    \item Our results show that PGMR drastically reduces URI hallucinations while significantly improving performance across LLMs and datasets, with most cases showing an almost complete elimination of hallucinated URIs.
    \item We demonstrate that a retrieval confidence threshold enables PGMR to correctly refuse to answer queries based on incomplete knowledge, thereby ensuring safety and robustness.
    \item PGMR exhibits minimal performance degradation (an approximate 4\% SQM drop) when the memory is subjected to a ninefold increase in memory size and irrelevant data.
\end{enumerate}

\section{Related Work}
\label{sec:related_work}

\noindent \textbf{SPARQL Query Generation using LLMs.} Even though various models have been proposed for the SPARQL query generation task recently, they suffer from one key problem: they require tagged question data \citep{banerjee, reyd2023, karou2023, qi2024enhancing}. In tagged data, the input to the model contains the natural language question, along with the URIs and their labels, necessary to formulate the corresponding SPARQL query. 
This requirement presents a challenge due to the high costs and labor-intensive nature of tagging, making it unsuitable for practical, real-world use cases. Therefore, our study emphasizes and assesses the untagged versions of the datasets.

\citet{reyd2023} and \citet{karou2023} propose a copy mechanism for copying URIs from the tagged question while generating the SPARQL query using LLMs. Additionally, they investigate the models' performance in dealing with novel question-query structures and unknown URIs. We incorporate this evaluation method to measure the out-of-distribution robustness of our proposed approach.
\citet{banerjee} propose a pointer generator network-based approach for SPARQL query generation over tagged data, utilizing BERT \citep{bert} and a finetuned T5. Their approach generates multiple SPARQL queries and selects the first query that executes and fetches an answer from the knowledge base as the predicted SPARQL query. This evaluation strategy contrasts with our approach of generating only one SPARQL query per question, which is more efficient. 
\citet{qi2024enhancing} employ a pre-training method that integrates the proposed Triple Structure Correction (TSC) strategy. This technique, which requires tagged data, involves randomly interchanging the positions of the subject, predicate, and object in SPARQL query triples with a certain probability, alongside a Masked Language Modeling (MLM) objective in a multi-task learning setup, before proceeding with fine-tuning on the downstream tagged dataset.
\citet{zahera2024cot} explore chain-of-thought prompting \citep{cot_paper} for SPARQL query generation.
RAG for SPARQL query generation has also been explored in recent works \citep{emonet2024llm}. Our comparative baseline adopts a comparable RAG approach.

\noindent \textbf{Mitigating hallucinations in LLMs.} Despite their widespread adoption, LLMs are vulnerable to generating factually incorrect information through hallucinations, which affects their reliability in real-world applications \citep{hallucination_survey_1, hallucination_survey_2, huang2021factual}. This challenge has led to significant research aimed at hallucination detection and mitigation \citep{lin2024towards, varshney2023stitch, dhuliawala2023chain, chern2023factool, li2023halueval}.
\citet{varshney2023stitch} identify potential hallucination candidates using the LLM’s logit outputs, validating their correctness, addressing any detected hallucinations, and then proceeding with the generation process.
In contrast, PGMR lets the LLM hallucinate between special tags and corrects these hallucinations post-generation using memory retrieval.
The Chain-of-Verification (CoVe) \citep{dhuliawala2023chain} method involves the model first producing an initial draft, then formulating verification questions to validate it by independently answering these questions, and generating the final, verified output.
Our method instead relies on a non-parametric memory to verify and include factually correct information from a KG.



\noindent \textbf{External memory augmented LLMs.} Earlier studies have investigated methods to augment LLMs by retrieving documents from external memory and integrating them into the contextual framework of tasks to provide relevant information.
According to \citet{retllm} and \citet{memllm}, the lack of a specialized memory unit in current LLMs constrains their ability to store and retrieve knowledge relevant to tasks explicitly. This observation supports the central premise of our study. They suggest enhancing LLMs with an API-based memory unit for read-write operations to address this limitation.
Interactions with external memory can be implemented using natural language interfaces \citep{park2023generative, zhou2023recurrentgpt} or formal mechanisms, such as standardized APIs, that allow the model to parse and execute commands \citep{toolformer, hu2023chatdb, retllm, memllm}.
Our approach is akin to the use of a formal mechanism for memory retrieval through our intermediate query format (as described in \refsec{sec:transform}).

\section{Preliminaries}
\label{sec:prelims}


\subsection{Uniform Resource Identifiers (URIs)}
\label{sec:uri}
In a large KG like Wikidata, each entity, relation, and property is assigned a Uniform Resource Identifier (URI), which is associated with a corresponding natural language URI label and other relevant metadata. For example, for an entity with the URI label "The Truman Show," the associated Wikidata URI is "Q214801." There are two types of URIs in the Wikidata KG, namely, "Q-ids," which correspond to the entities (e.g., "Q76" refers to the entity "Barack Obama"), and "P-ids," which correspond to relations, properties, and classes in the KG (e.g., "P26" refers to the relation "spouse").

\subsection{SPARQL Queries}
\label{sec:sparql_queries}
SPARQL is the standard declarative query language for Resource Description Framework (RDF) based KGs, such as Wikidata. It enables data retrieval by matching triple patterns corresponding to subject-predicate-object relationships. 
For example, a natural language question like \textit{"For which film was Sergei Eisenstein the film editor?"} can be translated into the following SPARQL query - "\textit{select distinct ?sbj where \{ ?sbj wdt:p1040 wd:q8003 . ?sbj wdt:p31 wd:q11424 \}}," where p1040, q8003, p31, and q11424 are URIs representing entities, relations, and properties from a KG like Wikidata\footnote{\url{https://www.wikidata.org}}.
However, authoring SPARQL queries is challenging for both non-experts and LLMs, as it requires precise knowledge of the KG's schema and opaque URIs.
Generating Wikidata's non-semantic URIs (e.g., a letter followed by arbitrary digits) forces an LLM to rely solely on its parametric memory to recall identifier mappings.




\begin{figure*}
     \centering
    \includegraphics[width=\textwidth]{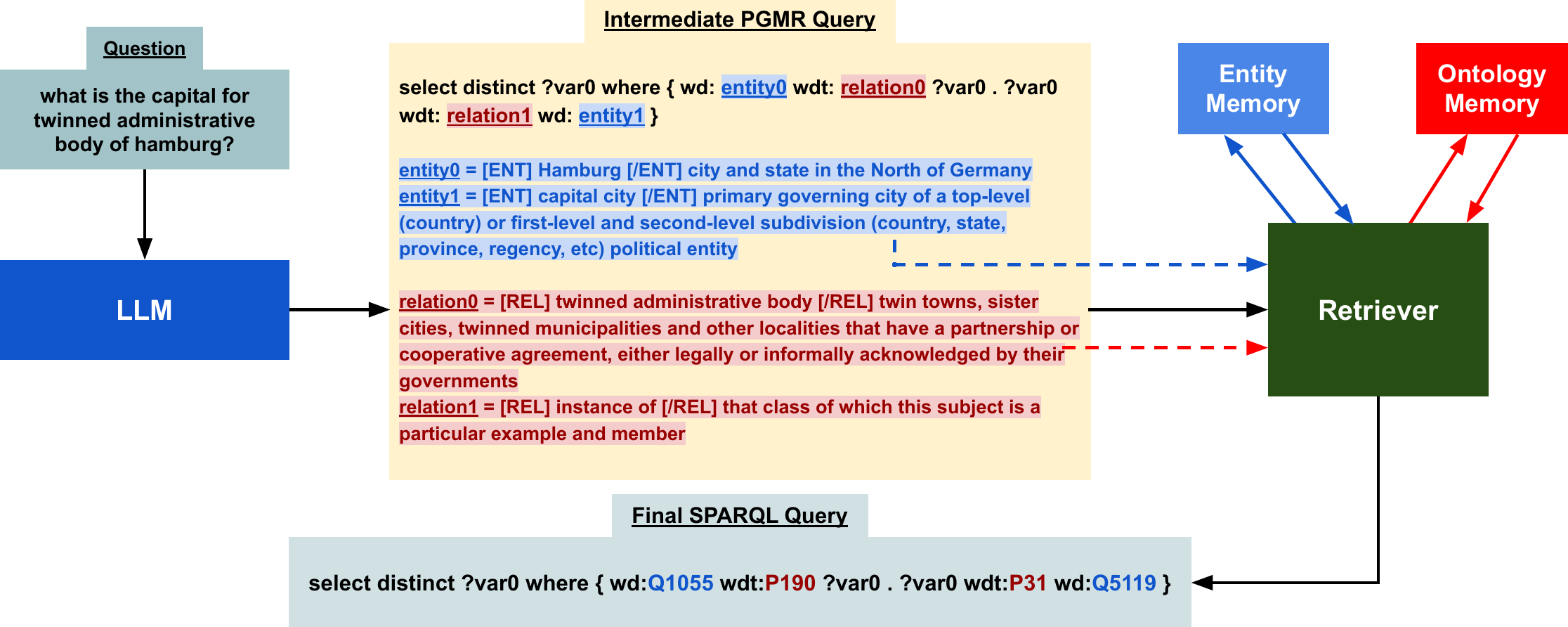}
    \caption{As discussed in \refsec{sec:pgmr}, PGMR first employs an LLM to produce an intermediate SPARQL-PGMR query, where all entities and relations are replaced by placeholders which are mapped to natural language labels and definitions, rather than formal URIs (\refsec{sec:transform}). The retriever (\refsec{sec:retriever}) subsequently fetches and replaces these labels with the most similar URIs from the relevant non-parametric memory to create the final SPARQL query.}
    \label{fig:genR}
\end{figure*}

\section{Post-Generation Memory Retrieval (PGMR)}
\label{sec:pgmr}

In this section, we introduce PGMR (“Post-Generation Memory Retrieval”), a novel method for SPARQL query generation utilizing LLMs. In PGMR, we initially convert SPARQL queries into an intermediate SPARQL-PGMR query format.
This intermediate query representation preserves the syntax of SPARQL while replacing URIs with placeholders that explicitly encode entities and relations in natural language. Specifically, entities are denoted as entityX = [ENT] Label [/ENT] Description, where Label provides the canonical surface form of the entity and Description gives a concise disambiguating context. Relations follow a parallel scheme, expressed as relationY = [REL] Label [/REL] Description, where the description clarifies the semantics of the property and distinguishes it from closely related relations.
These intermediate queries are later grounded in KG URIs using a retriever (discussed in \refsec{sec:retriever}) to produce the final SPARQL query.


As seen in \reffig{fig:genR}, a SPARQL-PGMR intermediate query thus consists of two components: (1) a SPARQL query block that maintains the original query structure but replaces URIs with entity and relation placeholders, and (2) a mapping block that defines the corresponding natural language labels and disambiguating descriptions for each placeholder. This design ensures syntactic fidelity to SPARQL while maintaining human readability, thereby enabling LLMs to generate valid queries that can be reliably grounded to knowledge graph URIs in a post-processing step.
This representation achieves two goals: (i) it constrains generation to a valid SPARQL structure, ensuring compatibility with downstream execution, and (ii) it provides natural language grounding for entities and relations, enabling post-generation alignment with the KG using a non-parametric memory of URIs and a retriever. By decoupling the SPARQL structural generation from the disambiguation of identifiers, SPARQL-PGMR supports accurate query generation and facilitates transparent inspection of model outputs.
Unlike LLM-based direct SPARQL query generation, which may produce URIs not present in the KG, PGMR's approach of conducting memory retrieval post-LLM generation greatly mitigates the risk of hallucinating non-existent KG URIs, as discussed in \refsec{sec:results}.

\begin{table*}[]
\centering
\resizebox{2\columnwidth}{!}{%
\begin{tabular}{|c|c|c|c|c|c|c|c|}
\hline

\textbf{Model} & \textbf{Type} & \textbf{F1 $\uparrow$} & \textbf{SQM $\uparrow$} & \textbf{BLEU $\uparrow$} & \textbf{Qid EM $\uparrow$} & \textbf{Pid EM $\uparrow$} & \textbf{URI Hallucination $\downarrow$} \\ \hline
\textbf{T5-Small} & FT & 44.84 & 37.82 \% & 81.54 & 42.93 \% & 89.69 \% & 30.02 \% \\ \hline
\textbf{T5-Small RAG} & FT &  67.30 & 67.03 \% & 89.96 & 73.68 \% & 95.48 \% & 14.70 \% \\ \hline
\textbf{\begin{tabular}[c]{@{}c@{}}T5-Small PGMR\\ (ours)\end{tabular}} & FT &  \textbf{86.75}  & \textbf{86.37 \%} & 94.63 & 93.17 \% & 97.05 \% & \textbf{0.0 \%} \\ \hline \hline

\textbf{Llama 3.1 8B} & PEFT & 19.89 & 0.75 \% & 66.74 & 1.52 \% & 61.23 \% & 89.84 \%  \\ \hline
\textbf{Llama 3.1 8B RAG} & PEFT &  81.66  & 81.37 \% & 94.65 & 90.19 \%  & 91.83 \% & 7.23 \% \\ \hline
\textbf{\begin{tabular}[c]{@{}c@{}}Llama 3.1 8B PGMR\\ (ours)\end{tabular}} & PEFT &  \textbf{88.73} & \textbf{87.84 \%} & 95.88 & 94.26 \% & 93.17 \% & \textbf{0.0 \%} \\ \hline \hline

\textbf{GPT 4o} & 25-shot &  17.27  & 2.82 \% & 56.65 & 7.04 \% & 50.11 \% & 81.02 \% \\ \hline
\textbf{GPT 4o RAG} & 25-shot &  58.85  & 27.21 \% & 69.79 & 82.27 \% & 65.86 \% & 4.88 \% \\ \hline
\textbf{\begin{tabular}[c]{@{}c@{}}GPT 4o PGMR\\ (ours)\end{tabular}} & 25-shot &  \textbf{62.23}  & \textbf{51.24 \%} & 77.49 & 79.12 \% & 68.10 \% & \textbf{ 0.01 \%} \\ \hline

\end{tabular}%
}
\caption{\textbf{Results on LCQUAD 2.0 (original split):} Across different LLMs, PGMR demonstrates a substantial reduction in hallucinations while achieving significantly higher SQM and answer-level F1 scores than direct SPARQL generation and RAG, as discussed in \refsec{sec:results}.}
\label{tbl:t5_lc2_og_results}
\end{table*}




\subsection{Training Data Transformation}
\label{sec:transform}
To train models capable of generating SPARQL-PGMR queries, we design a systematic transformation process that converts standard question-SPARQL query pairs into the PGMR representation. The transformation consists of three stages: entity replacement, relation replacement, and mapping construction.
In the first stage, all entity URIs (e.g., Q937) appearing in the query are replaced with symbolic placeholders of the form entityX. A mapping entry is then created for each placeholder, expressed as entityX = [ENT] Label [/ENT] Description, where the label corresponds to the canonical surface form and the description provides a short disambiguating definition extracted from the KG (e.g., [ENT] Albert Einstein [/ENT] German-born theoretical physicist, developer of the theory of relativity, Nobel Prize laureate (1921)).
The second stage applies the same procedure to relation and property URIs (e.g., P35), replacing them with placeholders such as relationY and constructing mappings of the form relationY = [REL] Label [/REL] Description. The description obtained from the KG provides the necessary semantic information for relation disambiguation (e.g., [REL] head of state [/REL] official with the highest formal authority in a country/state).
Finally, a mapping block is appended to each transformed query, listing all entity and relation definitions.
In this study, we obtain labels and descriptions from Wikidata using its official Python library\footnote{\url{https://pypi.org/project/Wikidata/}}.

\subsection{Memory Retriever}
\label{sec:retriever}
\noindent \textbf{Memories:} We utilize two memory modules (one for entities and another for relations) designed as a dictionary that pairs LLM encoder-generated embeddings of URI labels and their descriptions as keys with the corresponding URIs as values. The entity memory module includes such embeddings and URIs for all entity URIs in the dataset. Similarly, the ontology memory module includes these for all the relation URIs in the dataset.


\noindent \textbf{Retriever:} Given a natural language input such as "[ENT] Albert Einstein [/ENT] German-born theoretical physicist, developer of the theory of relativity, Nobel Prize laureate (1921)," our memory retriever locates the URI whose label and description are most similar to this text within the relevant memory.
The label is first parsed and matched with the URI labels in the memory.
If there is no direct match or multiple matches for the label, then the input text with the label and description is encoded into an embedding by an LLM encoder.
The retriever then uses FAISS \citep{faiss_lib,faiss_gpu} to retrieve the URI from the memory whose embedding is nearest to the input text embedding.
We use BGE \citep{chen2024bge} as the encoder for our retriever in this study.



\subsection{Setup}
\label{sec:inference}

\noindent \textbf{Training Setup:} Using the transformed training data from \refsec{sec:transform}, we finetune the LLM to generate intermediate queries given the questions.

\noindent \textbf{Few-shot Setup:} Drawing on the transformed training data, we apply few-shot prompting to the LLM, enabling it to generate intermediate queries.

\noindent \textbf{Inference:} During the inference phase, we generate a SPARQL-PGMR query for each question using the LLM. Each URI placeholder is then replaced by its corresponding URI through the retriever, thereby forming the final SPARQL query.


\begin{table*}[]
\centering
\resizebox{2\columnwidth}{!}{%
\begin{tabular}{|c|c|c|c|c|c|c|c|}
\hline

\textbf{Model} & \textbf{Type} & \textbf{F1 $\uparrow$} & \textbf{SQM $\uparrow$} & \textbf{BLEU $\uparrow$} & \textbf{Qid EM $\uparrow$} & \textbf{Pid EM $\uparrow$} & \textbf{URI Hallucination $\downarrow$} \\ \hline
\textbf{T5-Small} & FT &   22.67  & 0.16 \% & 70.44 & 2.71 & 84.75 \% & 61.66 \% \\ \hline
\textbf{T5-Small RAG} & FT &   67.25   & 64.14 \% & 89.85  & 71.98 \% & 94.11 \% & 16.01 \% \\ \hline
\textbf{\begin{tabular}[c]{@{}c@{}}T5-Small PGMR\\ (ours)\end{tabular}} & FT &   \textbf{82.13}  & \textbf{81.57 \%} & 94.24 & 90.4 \% & 95.93 \% & \textbf{ 0.0 \%} \\ \hline \hline

\textbf{Llama 3.1 8B} & PEFT &   22.19   & 0.03 \% & 67.33 & 0.53 \% & 65.40 \% & 91.04 \% \\ \hline
\textbf{Llama 3.1 8B RAG} & PEFT &   82.35   & 81.24 \% & 94.95  & 89.67 \% & 91.10 \% & 7.71 \% \\ \hline
\textbf{\begin{tabular}[c]{@{}c@{}}Llama 3.1 8B PGMR\\ (ours)\end{tabular}} & PEFT &   \textbf{84.47}   & \textbf{83.33 \%} & 94.35 & 92.65 \% & 90.80 \% & \textbf{0.0 \%} \\ \hline \hline

\textbf{GPT 4o} & 25-shot &   17.27   & 1.45 \% & 55.39 & 4.07 \% & 48.92 \% & 85.87 \% \\ \hline
\textbf{GPT 4o RAG} & 25-shot &   53.79   & 23.65 \% & 64.28 & 83.46 \% & 63.25 \% & 3.47 \% \\ \hline
\textbf{\begin{tabular}[c]{@{}c@{}}GPT 4o PGMR\\ (ours)\end{tabular}} & 25-shot &   \textbf{69.05}   & \textbf{58.42 \%} & 82.77 & 86.14 \% & 73.10 \% & \textbf{ 0.0 \%} \\ \hline

\end{tabular}%
}
\caption{\textbf{Results on LCQUAD 2.0 (unknown URI split):} Even in an out-of-distribution setting, PGMR significantly mitigates hallucinations while enhancing SQM and answer-level F1 performance over the baselines across various LLMs and settings (see \refsec{sec:results}).}
\label{tbl:t5_lc2_unk_uri_results}
\end{table*}

\section{Experimental Setup}
\label{sec:experiments}


\subsection{Datasets}
\label{sec:data}
We evaluate our work on two widely used datasets for SPARQL query generation: LCQUAD 2.0 \citep{lcquad2} and QALD-10 \citep{qald10}.
Even though tagged versions of these datasets exist, where the entities and relations in the question are already linked and tagged with associated URIs from the KG, we use the untagged versions. Tagging is expensive, time-consuming, and impractical in the real world. 
SPARQL query generation becomes a much harder problem when using untagged data (see \refsec{sec:results}).

\noindent \textbf{LCQUAD 2.0} comprises a mix of simple and complex questions with associated SPARQL queries over Wikidata, crafted by human annotators from Amazon Mechanical Turk. The \textbf{original split} of LCQUAD 2.0 includes a broad dataset with 21k questions for training, 3k questions for validation, and 6k questions for testing. To specifically test how our models perform in the face of out-of-distribution URIs not seen before in training, we also construct an \textbf{"unknown URI" split} of LCQUAD 2.0 in line with \citet{reyd2023}. Every query in the test set of this version of the dataset includes at least one URI not seen during training, thus making accurate query generation a much more complex problem. The "unknown URI" split contains 24k train, 3k validation, and 3k test questions and associated queries.

\noindent \textbf{QALD-10} is a benchmarking dataset that is a part of the  Question Answering over Linked Data (QALD) challenge series for KGQA using SPARQL query generation. Owing to the complexity of SPARQL queries that involve various clauses and literals, QALD-10 is recognized as one of the most challenging and practically applicable datasets in the QALD challenge series. Although each question in QALD-10 is translated into eight different languages, we only consider the English versions of the questions for this analysis. The dataset consists of 412 training questions and 394 test questions, which is much smaller than LCQUAD 2.0 but contains more complex queries.
Given the limited availability of training data for QALD-10, we employ a two-step approach in our finetuning experiments: pre-training the LLM on LCQUAD 2.0 before finetuning it on QALD-10.


\begin{table*}[]
\centering
\resizebox{2\columnwidth}{!}{%
\begin{tabular}{|c|c|c|c|c|c|c|c|}
\hline

\textbf{Model} & \textbf{Type}  & \textbf{F1 $\uparrow$} & \textbf{SQM $\uparrow$} & \textbf{BLEU $\uparrow$} & \textbf{Qid EM $\uparrow$} & \textbf{Pid EM $\uparrow$} & \textbf{URI Hallucination $\downarrow$} \\ \hline
\textbf{T5-Small} & FT   &  11.30 & 5.07 \% & 34.37 & 11.42 \% & 25.38 \% & 66.50 \% \\ \hline
\textbf{T5-Small RAG} & FT   & \textbf{23.15}  & 15.98 \% & 43.47 & 45.68 \% & 27.66 \% & 4.82 \% \\ \hline
\textbf{\begin{tabular}[c]{@{}c@{}}T5-Small PGMR\\ (ours)\end{tabular}} & FT  &  22.12 & \textbf{16.75 \%} & 44.31 & 42.39 \% & 27.41 \% & \textbf{0.0 \%} \\ \hline \hline

\textbf{Llama 3.1 8B} & PEFT  & 3.93  & 0.25 \% & 24.35 & 1.27 \% & 22.33 \% & 89.09 \% \\ \hline
\textbf{Llama 3.1 8B RAG} & PEFT  & 35.46  & 23.35 \% & 45.91 & 60.15 \% & 36.55 \% & 8.12 \% \\ \hline
\textbf{\begin{tabular}[c]{@{}c@{}}Llama 3.1 8B PGMR\\ (ours)\end{tabular}} & PEFT   & \textbf{39.41}  & \textbf{25.63 \%} & 45.36 & 53.55 \% & 37.81 \% & \textbf{0.0 \%} \\ \hline \hline

\textbf{GPT 4o} & 25-shot   & 13.20  & 3.55 \% & 41.21 & 11.93 \% & 46.19 \% & 83.76 \% \\ \hline
\textbf{GPT 4o RAG} & 25-shot   & 46.52  & 28.68 \% & 51.66 & 74.62 \% & 53.55 \% & 11.42 \% \\ \hline
\textbf{\begin{tabular}[c]{@{}c@{}}GPT 4o PGMR\\ (ours)\end{tabular}} & 25-shot   & \textbf{51.19}  & \textbf{29.19 \%} & 54.95 & 65.48 \% & 46.19 \% & \textbf{0.0 \%} \\ \hline

\end{tabular}%
}
\caption{\textbf{Results on QALD-10:} PGMR consistently minimizes hallucinations across LLMs while attaining significantly better SQM and answer-level F1 scores than direct SPARQL generation, even for the more complex QALD-10 dataset, as discussed in \refsec{sec:results}.}
\label{tbl:t5_qald_results}
\end{table*}



\subsection{Metrics}
\label{sec:metrics}

To measure the effectiveness of our models, we utilize the following three metrics: 

\noindent \textbf{F1 Score:}
We execute the gold and the generated SPARQL queries and then calculate the answer-level F1 score based on the responses.

\noindent \textbf{Semantic Query Match (SQM):} 
We introduce the SQM metric, which assesses semantic equivalence rather than literal string match. Recognizing that LLMs may produce valid queries with different variable names than the reference (e.g., ?value vs. ?uri), we normalize all variables to a canonical representation (e.g., ?var0, ?var1), a technique consistent with the evaluation framework of \citet{qi2024enhancing}.
Additionally, since the ordering of triples within a WHERE clause and of entities within a triple does not alter the query's logical semantics (queries would result in the same outcome when executed), our comparison is insensitive to their permutation.
Once the variables are normalized and the WHERE clause is compared in a permutation-invariant manner, the remainder of the query is subjected to an exact-match comparison.

\noindent \textbf{BLEU} score \citep{bleu} is a popular neural machine translation metric that assesses the predicted query against the gold standard query by comparing tokens.  

\noindent \textbf{Qid-EM and Pid-EM:} To assess the Qid exact match (EM) score, we directly evaluate whether the set of all entity URIs in the predicted query matches exactly with those in the reference query. Pid EM does the same for relations and properties.

\noindent \textbf{URI Hallucination:}  We define a new metric, called URI Hallucination, which measures the percentage of queries where at least one URI is hallucinated. A URI is considered hallucinated if it does not exist in the knowledge base memory.

\noindent \textbf{Refusal Accuracy} is defined as the success rate with which the model identifies and correctly signals a lack of the requisite internal knowledge, thereby declining to answer the unsupported query (discussed further in \refsec{sec:analysis}).

\subsection{Baselines}
\label{sec:baselines}


\noindent \textbf{Direct Generation} refers to the standard scenario where the LLM directly generates the SPARQL query given a question.

\noindent \textbf{RAG:} We additionally compare our results against the RAG paradigm, a strong baseline, wherein the question is initially processed by a retriever that retrieves $k$ URIs from the URI memory whose labels and descriptions most closely match the question text in latent space, as described in \refsec{sec:rag}.

\noindent \textbf{PGMR:} As described in \refsec{sec:pgmr}, we implement the proposed PGMR, where an LLM first generates intermediate queries, followed by a retrieval step to translate them into SPARQL queries.

\subsubsection{Language Models}
\label{sec:llms_used}
Our evaluation incorporates a diverse range of LLMs spanning several orders of magnitude in parameter size, including models from the smaller T5 architecture \citep{t5} (fine-tuning - FT) up to Llama 3.1 8B Instruct (parameter-efficient fine-tuning - PEFT) \citep{llama3,peft} and the foundational GPT-4o (25-shot prompting) \citep{gpt4}.


\section{Results}
\label{sec:results}


\noindent \textbf{PGMR enhances the quality of SPARQL queries.}
Across all assessed LLMs, PGMR consistently enhances the accuracy and structure of SPARQL queries on both LCQUAD 2.0 and QALD-10 datasets. As seen in Tables \ref{tbl:t5_lc2_og_results}, \ref{tbl:t5_lc2_unk_uri_results}, and \ref{tbl:t5_qald_results}, PGMR improves the SQM across all baselines, LLMs, and datasets.
On LCQUAD2 (original split), PGMR yields a performance gain of nearly 61.35 \% over direct generation and 16.61 \% over the strong RAG baseline on average across LLMs.
Notably, this advantage continues on the out-of-distribution "unknown URI" split of LCQUAD2, with PGMR improving 73.89 \% over direct generation and 18.09 \% over RAG across LLMs, showing its robustness to shifts in data distribution.
Even on the more challenging QALD-10 dataset, PGMR outperforms both direct generation and RAG across LLMs.

\noindent \textbf{PGMR nearly eliminates hallucinations.}
From Tables \ref{tbl:t5_lc2_og_results}, \ref{tbl:t5_lc2_unk_uri_results}, and \ref{tbl:t5_qald_results}, we see that direct generation hallucinates URIs 75.42 \% of the time on average across datasets and models. While RAG reduces these hallucinations to 8.71 \% on average, PGMR achieves a near-complete elimination of the URI hallucination problem regardless of the LLM, dataset, and data distribution used.

Our results confirm that PGMR not only minimizes hallucinations to a great extent but also improves the accuracy and reliability of SPARQL query generation across various LLMs, datasets, and data distributions.


\begin{figure}[ht]
    \centering
    \begin{subfigure}[b]{0.48\textwidth}
        \centering
        \includegraphics[width=\linewidth]{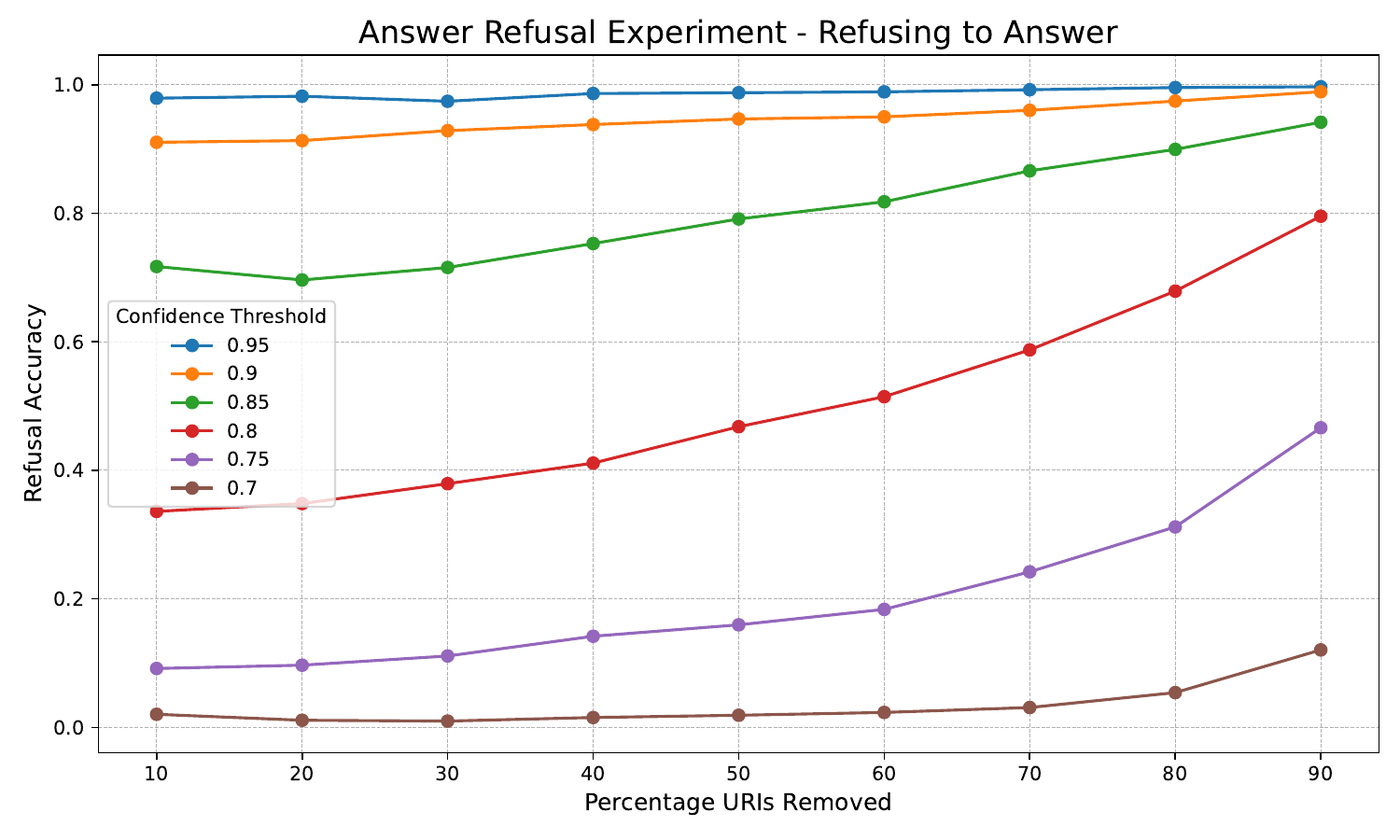}
        \caption{}
        \label{fig:a}
    \end{subfigure}
    \hfill
    \begin{subfigure}[b]{0.48\textwidth}
        \centering
        \includegraphics[width=\linewidth]{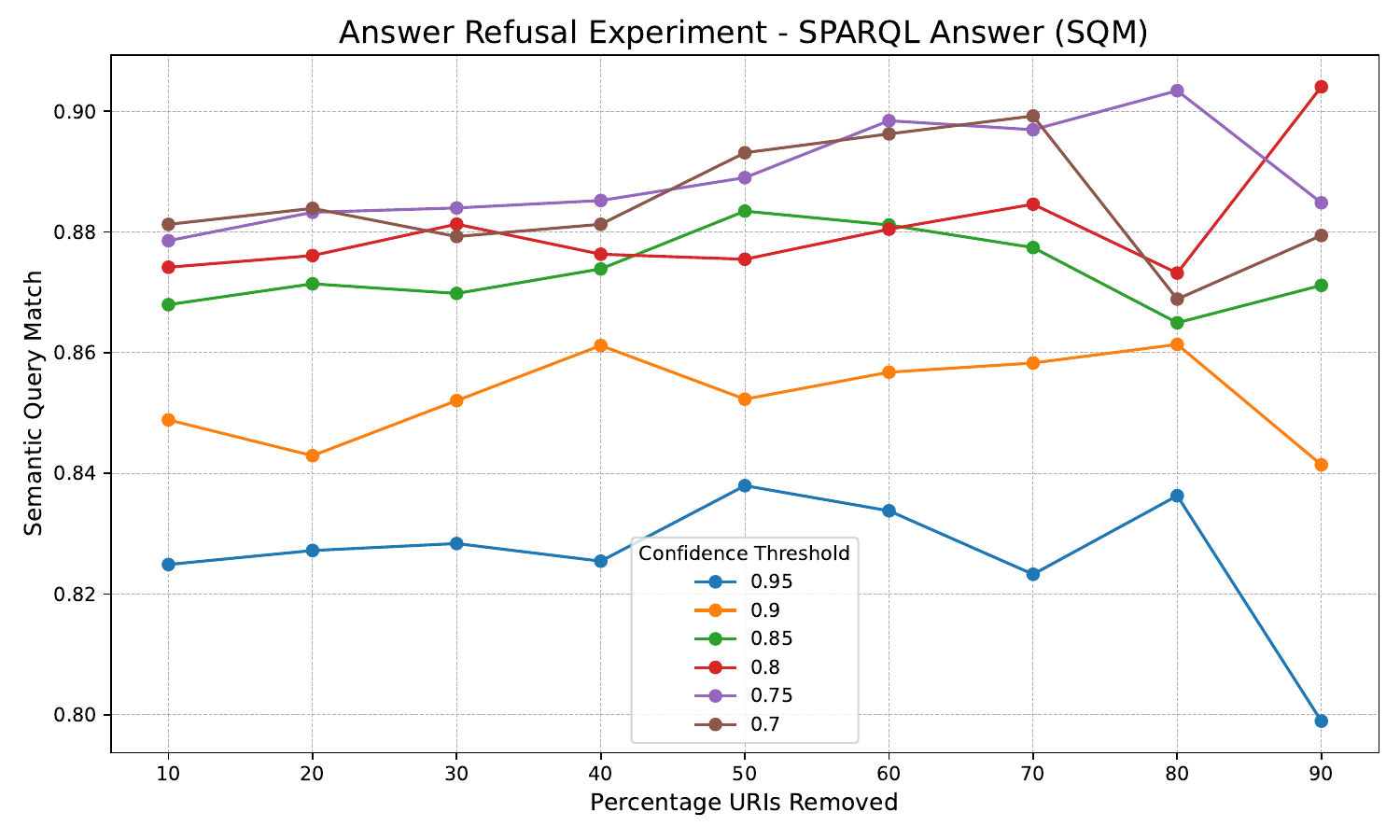}
        \caption{}
        \label{fig:b}
    \end{subfigure}
    \caption{For LCQUAD2, \reffig{fig:a} shows the accuracy of Llama 3.1 8B PGMR on the unanswerable set as a function of the percentage of URIs removed across confidence thresholds, while \reffig{fig:b} shows the SQM on the answerable set (described in \refsec{sec:analysis}).}
    \label{fig:combined}
\end{figure}


\section{Analysis and discussion}
\label{sec:analysis}

\noindent \textbf{Answer Refusal Experiment.}
A model's ability to abstain from responding when it lacks the required information is a key determinant of its safety and trustworthiness. This mechanism directly counters model hallucination by signaling uncertainty instead of fabricating content. Such a conservative approach is indispensable in critical applications where a confidently false answer has severe consequences.
We assess such robustness by randomly ablating a percentage of entity URIs from the memory, rendering dependent test queries unanswerable. We then divide the set, calculating SQM on the answerable queries and refusal accuracy on the unanswerable queries.
The PGMR model architecture is inherently equipped to handle this uncertainty without additional training, thanks to the implementation of a confidence threshold applied to the retrieval similarity score. Specifically, the model executes a refusal action if the similarity score of the top-ranked retrieved URI falls below this predetermined threshold for any URI required in the generated query.
Figure \ref{fig:a} demonstrates a clear positive correlation between the confidence threshold and refusal accuracy for Llama 3.1 8B PGMR. Conversely, as seen in \reffig{fig:b}, this increase in threshold leads to a corresponding decline in SQM on the answerable subset, illustrating a direct trade-off in performance. Our analysis indicates that the optimal threshold value for balancing these two metrics is approximately 0.85.
Furthermore, this confidence threshold can potentially be learned by adding a small neural network with a sigmoid head to the retriever's encoder, but we leave this exploration for future work.


\begin{figure}[ht]
    \centering
    \includegraphics[width=\linewidth]{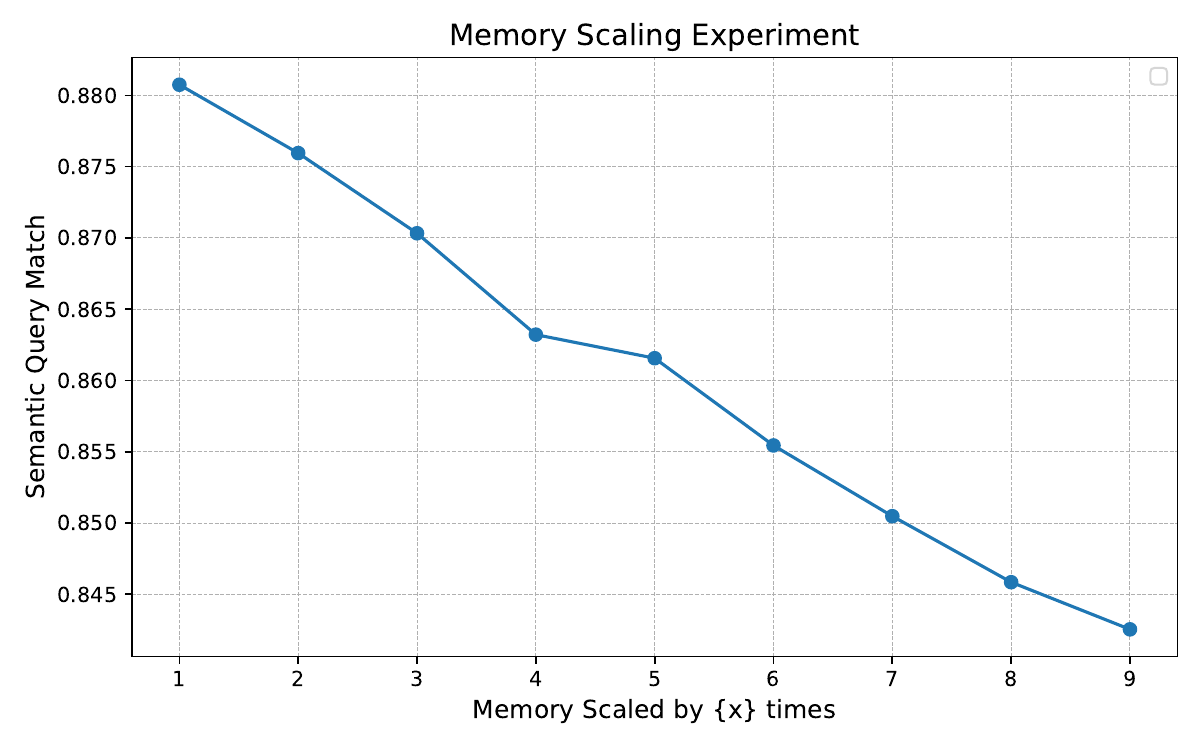} 
    \caption{Llama 3.1 8B PGMR maintained robust performance on LCQUAD2, exhibiting only an approximate 4\% drop in SQM despite a 9× memory size increase. This demonstrates the retriever's resilience to memory scaling and noise (discussed in \refsec{sec:analysis}).}
    \label{fig:memory_scaling}
\end{figure}

\noindent \textbf{Memory Scaling Experiment.}
We investigate the scaling robustness of the PGMR framework, manipulating the entity memory size from a factor of 2 to 9 relative to the original set.
This procedure involves augmenting the memory with irrelevant entities sourced from the Wikidata KG.
These entities serve as distractors during the retriever's SPARQL URI resolution.
From \reffig{fig:memory_scaling}, we see that Llama 3.1 8B PGMR exhibits minimal performance degradation of approximately 4\% on LCQUAD2, even at 9 times the memory size, validating the retriever's robustness to expanded, noisy memory.

\noindent \textbf{Inference Latency Analysis.}
Under equivalent constraints for maximum input and output token lengths, the inference latency of the proposed methods can be characterized by the base generation time ($T_{dg}$) and the retrieval overhead ($T_{ret}$). For a model such as Llama 3.1 8B, $T_{dg}$ is approximately 0.9 seconds per response. In contrast, the retrieval component is highly efficient, requiring only 0.0016 seconds ($T_{ret}$) and a modest 600MB of additional GPU memory. Consequently, while Direct Generation time is simply $T_{dg}$, RAG latency is defined as $T_{rag} = T_{dg} + T_{ret}$. The latency for PGMR is expressed as $T_{pgmr} = T_{dg} + (k \cdot T_{ret})$, where $k$ represents the number of URIs identified within the generated intermediate query. Given that $T_{ret} \ll T_{dg}$, the computational overhead introduced by PGMR remains negligible relative to the base model's inference time.
\section{Conclusion}
\label{sec:conclusion}

We presented PGMR (Post-Generation Memory Retrieval), a modular framework that addresses URI hallucination in LLM-generated SPARQL queries by separating query logic generation from retrieval.
LLMs generate intermediate queries with URI placeholders and associated URI natural language mappings, which a non-parametric retriever then resolves into accurate URIs.
Despite its simplicity, PGMR consistently achieved superior semantic query match (SQM) and answer-level F1 scores across diverse LLMs and datasets, yielding a near-complete elimination of URI hallucinations.
Crucially, the system demonstrates exceptional robustness, both in its native ability to refuse to answer queries based on confidence thresholds and its resilience to memory scaling, maintaining high SQM with only minimal performance degradation even when memory was expanded nine-fold with distracting entities.

\section*{Limitations}

Even though PGMR exhibits markedly improved performance in generating correct URIs and reducing hallucinations, its effectiveness is tempered by the LLM's ability to construct SPARQL query structures accurately.
This is especially evident when looking at the results on the QALD-10 dataset (see \reftbl{tbl:t5_qald_results}), which features notably complex queries involving diverse clauses and literal expressions compared to LCQUAD 2.0.
This limitation may be addressed by using more powerful, large LLMs pre-trained specifically on SPARQL query data, combined with the PGMR generation paradigm, or by combining the advantages of RAG and PGMR, adding a retrieval step both before and after generation. However, we leave that for future work.

\section*{Acknowledgements}
\label{sec:acknowledgements}
We thank NSERC and Samsung for supporting this work and CIFAR for their support under the Canada CIFAR AI Chair program.
Additionally, we thank IVADO, as this project was undertaken thanks to funding from IVADO and the Canada First Research Excellence Fund.
We also thank Luis Lara for his help with running some of the experiments in the early part of this work.

\bibliography{references}

@inproceedings{bert,
  title={BERT: Pre-training of Deep Bidirectional Transformers for Language Understanding},
  author={Kenton, Jacob Devlin Ming-Wei Chang and Toutanova, Lee Kristina},
  booktitle={Proceedings of NAACL-HLT},
  pages={4171--4186},
  year={2019}
}

@article{t5,
  title={Exploring the limits of transfer learning with a unified text-to-text transformer},
  author={Raffel, Colin and Shazeer, Noam and Roberts, Adam and Lee, Katherine and Narang, Sharan and Matena, Michael and Zhou, Yanqi and Li, Wei and Liu, Peter J},
  journal={Journal of machine learning research},
  volume={21},
  number={140},
  pages={1--67},
  year={2020}
}

@article{faiss_gpu,
  title={Billion-scale similarity search with GPUs},
  author={Johnson, Jeff and Douze, Matthijs and J{\'e}gou, Herv{\'e}},
  journal={IEEE Transactions on Big Data},
  volume={7},
  number={3},
  pages={535--547},
  year={2019},
  publisher={IEEE}
}

@article{faiss_lib,
      title={The Faiss library},
      author={Matthijs Douze and Alexandr Guzhva and Chengqi Deng and Jeff Johnson and Gergely Szilvasy and Pierre-Emmanuel Mazaré and Maria Lomeli and Lucas Hosseini and Hervé Jégou},
      year={2024},
      eprint={2401.08281},
      archivePrefix={arXiv},
      primaryClass={cs.LG}
}

@inproceedings{knnlm,
  title={Generalization through Memorization: Nearest Neighbor Language Models},
  author={Khandelwal, Urvashi and Levy, Omer and Jurafsky, Dan and Zettlemoyer, Luke and Lewis, Mike},
  booktitle={International Conference on Learning Representations},
  year={2019}
}

@article{rag_og,
  title={Retrieval-augmented generation for knowledge-intensive nlp tasks},
  author={Lewis, Patrick and Perez, Ethan and Piktus, Aleksandra and Petroni, Fabio and Karpukhin, Vladimir and Goyal, Naman and K{\"u}ttler, Heinrich and Lewis, Mike and Yih, Wen-tau and Rockt{\"a}schel, Tim and others},
  journal={Advances in Neural Information Processing Systems},
  volume={33},
  pages={9459--9474},
  year={2020}
}

@article{Text2SPARQL,
  title={Leveraging small language models for Text2SPARQL tasks to improve the resilience of AI assistance},
  author={Brei, Felix and Frey, Johannes and Meyer, Lars-Peter},
  journal={arXiv preprint arXiv:2405.17076},
  year={2024}
}

@inproceedings{banerjee,
  title={Modern baselines for SPARQL semantic parsing},
  author={Banerjee, Debayan and Nair, Pranav Ajit and Kaur, Jivat Neet and Usbeck, Ricardo and Biemann, Chris},
  booktitle={Proceedings of the 45th International ACM SIGIR Conference on Research and Development in Information Retrieval},
  pages={2260--2265},
  year={2022}
}

@article{qi2024enhancing,
  title={Enhancing SPARQL Query Generation for Knowledge Base Question Answering Systems by Learning to Correct Triplets},
  author={Qi, Jiexing and Su, Chang and Guo, Zhixin and Wu, Lyuwen and Shen, Zanwei and Fu, Luoyi and Wang, Xinbing and Zhou, Chenghu},
  journal={Applied Sciences},
  volume={14},
  number={4},
  pages={1521},
  year={2024},
  publisher={MDPI}
}

@article{karou2023,
  title={A comprehensive evaluation of neural sparql query generation from natural language questions},
  author={Diallo, Papa Abdou Karim Karou and Reyd, Samuel and Zouaq, Amal},
  journal={IEEE Access},
  year={2024},
  publisher={IEEE}
}

@article{qald10,
  title={QALD-10--The 10th challenge on question answering over linked data},
  author={Usbeck, Ricardo and Yan, Xi and Perevalov, Aleksandr and Jiang, Longquan and Schulz, Julius and Kraft, Angelie and M{\"o}ller, Cedric and Huang, Junbo and Reineke, Jan and Ngonga Ngomo, Axel-Cyrille and others},
  journal={Semantic Web},
  number={Preprint},
  pages={1--15},
  year={2023},
  publisher={IOS Press}
}

@inproceedings{lcquad2,
  title={Lc-quad 2.0: A large dataset for complex question answering over wikidata and dbpedia},
  author={Dubey, Mohnish and Banerjee, Debayan and Abdelkawi, Abdelrahman and Lehmann, Jens},
  booktitle={The Semantic Web--ISWC 2019: 18th International Semantic Web Conference, Auckland, New Zealand, October 26--30, 2019, Proceedings, Part II 18},
  pages={69--78},
  year={2019},
  organization={Springer}
}

@inproceedings{bleu,
  title={Bleu: a method for automatic evaluation of machine translation},
  author={Papineni, Kishore and Roukos, Salim and Ward, Todd and Zhu, Wei-Jing},
  booktitle={Proceedings of the 40th annual meeting of the Association for Computational Linguistics},
  pages={311--318},
  year={2002}
}

@inproceedings{kgqa_survey,
  title={A Survey on Complex Knowledge Base Question Answering: Methods, Challenges and Solutions},
  author={Lan, Yunshi and He, Gaole and Jiang, Jinhao and Jiang, Jing and Zhao, Wayne Xin and Wen, Ji-Rong},
  booktitle={Proceedings of the Thirtieth International Joint Conference on Artificial Intelligence},
  year={2021},
  organization={International Joint Conferences on Artificial Intelligence Organization}
}

@article{kg_survey,
  title={Knowledge graphs: Opportunities and challenges},
  author={Peng, Ciyuan and Xia, Feng and Naseriparsa, Mehdi and Osborne, Francesco},
  journal={Artificial Intelligence Review},
  volume={56},
  number={11},
  pages={13071--13102},
  year={2023},
  publisher={Springer}
}

@inproceedings{saxena2020improving,
  title={Improving multi-hop question answering over knowledge graphs using knowledge base embeddings},
  author={Saxena, Apoorv and Tripathi, Aditay and Talukdar, Partha},
  booktitle={Proceedings of the 58th annual meeting of the association for computational linguistics},
  pages={4498--4507},
  year={2020}
}

@inproceedings{reyd2023,
  title={Assessing the Generalization Capabilities of Neural Machine Translation Models for SPARQL Query Generation},
  author={Reyd, Samuel and Zouaq, Amal},
  booktitle={International Semantic Web Conference},
  pages={484--501},
  year={2023},
  organization={Springer}
}

@article{retllm,
  title={Ret-llm: Towards a general read-write memory for large language models},
  author={Modarressi, Ali and Imani, Ayyoob and Fayyaz, Mohsen and Sch{\"u}tze, Hinrich},
  journal={arXiv preprint arXiv:2305.14322},
  year={2023}
}

@article{memllm,
  title={MemLLM: Finetuning LLMs to Use An Explicit Read-Write Memory},
  author={Modarressi, Ali and K{\"o}ksal, Abdullatif and Imani, Ayyoob and Fayyaz, Mohsen and Sch{\"u}tze, Hinrich},
  journal={arXiv preprint arXiv:2404.11672},
  year={2024}
}

@inproceedings{park2023generative,
  title={Generative agents: Interactive simulacra of human behavior},
  author={Park, Joon Sung and O'Brien, Joseph and Cai, Carrie Jun and Morris, Meredith Ringel and Liang, Percy and Bernstein, Michael S},
  booktitle={Proceedings of the 36th Annual ACM Symposium on User Interface Software and Technology},
  pages={1--22},
  year={2023}
}

@article{zhou2023recurrentgpt,
  title={Recurrentgpt: Interactive generation of (arbitrarily) long text},
  author={Zhou, Wangchunshu and Jiang, Yuchen Eleanor and Cui, Peng and Wang, Tiannan and Xiao, Zhenxin and Hou, Yifan and Cotterell, Ryan and Sachan, Mrinmaya},
  journal={arXiv preprint arXiv:2305.13304},
  year={2023}
}

@article{toolformer,
  title={Toolformer: Language models can teach themselves to use tools},
  author={Schick, Timo and Dwivedi-Yu, Jane and Dess{\`\i}, Roberto and Raileanu, Roberta and Lomeli, Maria and Hambro, Eric and Zettlemoyer, Luke and Cancedda, Nicola and Scialom, Thomas},
  journal={Advances in Neural Information Processing Systems},
  volume={36},
  year={2024}
}

@article{hu2023chatdb,
  title={Chatdb: Augmenting llms with databases as their symbolic memory},
  author={Hu, Chenxu and Fu, Jie and Du, Chenzhuang and Luo, Simian and Zhao, Junbo and Zhao, Hang},
  journal={arXiv preprint arXiv:2306.03901},
  year={2023}
}

@inproceedings{guu2020retrieval,
  title={Retrieval augmented language model pre-training},
  author={Guu, Kelvin and Lee, Kenton and Tung, Zora and Pasupat, Panupong and Chang, Mingwei},
  booktitle={International conference on machine learning},
  pages={3929--3938},
  year={2020},
  organization={PMLR}
}

@article{lewis2020retrieval,
  title={Retrieval-augmented generation for knowledge-intensive nlp tasks},
  author={Lewis, Patrick and Perez, Ethan and Piktus, Aleksandra and Petroni, Fabio and Karpukhin, Vladimir and Goyal, Naman and K{\"u}ttler, Heinrich and Lewis, Mike and Yih, Wen-tau and Rockt{\"a}schel, Tim and others},
  journal={Advances in Neural Information Processing Systems},
  volume={33},
  pages={9459--9474},
  year={2020}
}

@article{hallucination_survey_1,
  title={A survey on hallucination in large language models: Principles, taxonomy, challenges, and open questions},
  author={Huang, Lei and Yu, Weijiang and Ma, Weitao and Zhong, Weihong and Feng, Zhangyin and Wang, Haotian and Chen, Qianglong and Peng, Weihua and Feng, Xiaocheng and Qin, Bing and others},
  journal={arXiv preprint arXiv:2311.05232},
  year={2023}
}

@article{hallucination_survey_2,
  title={Survey of hallucination in natural language generation},
  author={Ji, Ziwei and Lee, Nayeon and Frieske, Rita and Yu, Tiezheng and Su, Dan and Xu, Yan and Ishii, Etsuko and Bang, Ye Jin and Madotto, Andrea and Fung, Pascale},
  journal={ACM Computing Surveys},
  volume={55},
  number={12},
  pages={1--38},
  year={2023},
  publisher={ACM New York, NY}
}

@article{huang2021factual,
  title={The factual inconsistency problem in abstractive text summarization: A survey},
  author={Huang, Yichong and Feng, Xiachong and Feng, Xiaocheng and Qin, Bing},
  journal={arXiv preprint arXiv:2104.14839},
  year={2021}
}

@article{chern2023factool,
  title={FacTool: Factuality Detection in Generative AI--A Tool Augmented Framework for Multi-Task and Multi-Domain Scenarios},
  author={Chern, I and Chern, Steffi and Chen, Shiqi and Yuan, Weizhe and Feng, Kehua and Zhou, Chunting and He, Junxian and Neubig, Graham and Liu, Pengfei and others},
  journal={arXiv preprint arXiv:2307.13528},
  year={2023}
}

@article{dhuliawala2023chain,
  title={Chain-of-verification reduces hallucination in large language models},
  author={Dhuliawala, Shehzaad and Komeili, Mojtaba and Xu, Jing and Raileanu, Roberta and Li, Xian and Celikyilmaz, Asli and Weston, Jason},
  journal={arXiv preprint arXiv:2309.11495},
  year={2023}
}

@inproceedings{li2023halueval,
  title={HaluEval: A Large-Scale Hallucination Evaluation Benchmark for Large Language Models},
  author={Li, Junyi and Cheng, Xiaoxue and Zhao, Wayne Xin and Nie, Jian-Yun and Wen, Ji-Rong},
  booktitle={Proceedings of the 2023 Conference on Empirical Methods in Natural Language Processing},
  pages={6449--6464},
  year={2023}
}

@article{varshney2023stitch,
  title={A stitch in time saves nine: Detecting and mitigating hallucinations of llms by validating low-confidence generation},
  author={Varshney, Neeraj and Yao, Wenlin and Zhang, Hongming and Chen, Jianshu and Yu, Dong},
  journal={arXiv preprint arXiv:2307.03987},
  year={2023}
}

@article{lin2024towards,
  title={Towards trustworthy LLMs: a review on debiasing and dehallucinating in large language models},
  author={Lin, Zichao and Guan, Shuyan and Zhang, Wending and Zhang, Huiyan and Li, Yugang and Zhang, Huaping},
  journal={Artificial Intelligence Review},
  volume={57},
  number={9},
  pages={243},
  year={2024},
  publisher={Springer}
}

@inproceedings{barnett2024seven,
  title={Seven failure points when engineering a retrieval augmented generation system},
  author={Barnett, Scott and Kurniawan, Stefanus and Thudumu, Srikanth and Brannelly, Zach and Abdelrazek, Mohamed},
  booktitle={Proceedings of the IEEE/ACM 3rd International Conference on AI Engineering-Software Engineering for AI},
  pages={194--199},
  year={2024}
}

@article{meyer2024assessing,
  title={Assessing SPARQL capabilities of Large Language Models},
  author={Meyer, Lars-Peter and Frey, Johannes and Brei, Felix and Arndt, Natanael},
  journal={arXiv preprint arXiv:2409.05925},
  year={2024}
}

@article{chen2024bge,
  title={Bge m3-embedding: Multi-lingual, multi-functionality, multi-granularity text embeddings through self-knowledge distillation},
  author={Chen, Jianlv and Xiao, Shitao and Zhang, Peitian and Luo, Kun and Lian, Defu and Liu, Zheng},
  journal={arXiv preprint arXiv:2402.03216},
  year={2024}
}

@article{llama3,
  title={The llama 3 herd of models},
  author={Dubey, Abhimanyu and Jauhri, Abhinav and Pandey, Abhinav and Kadian, Abhishek and Al-Dahle, Ahmad and Letman, Aiesha and Mathur, Akhil and Schelten, Alan and Yang, Amy and Fan, Angela and others},
  journal={arXiv preprint arXiv:2407.21783},
  year={2024}
}

@article{gpt4,
  title={Gpt-4 technical report},
  author={Achiam, Josh and Adler, Steven and Agarwal, Sandhini and Ahmad, Lama and Akkaya, Ilge and Aleman, Florencia Leoni and Almeida, Diogo and Altenschmidt, Janko and Altman, Sam and Anadkat, Shyamal and others},
  journal={arXiv preprint arXiv:2303.08774},
  year={2023}
}

@Misc{peft,
  title =        {{PEFT}: State-of-the-art Parameter-Efficient Fine-Tuning methods},
  author =       {Sourab Mangrulkar and Sylvain Gugger and Lysandre Debut and Younes Belkada and Sayak Paul and Benjamin Bossan},
  howpublished = {\url{https://github.com/huggingface/peft}},
  year =         {2022}
}

@article{emonet2024llm,
  title={Llm-based sparql query generation from natural language over federated knowledge graphs},
  author={Emonet, Vincent and Bolleman, Jerven and Duvaud, Severine and de Farias, Tarcisio Mendes and Sima, Ana Claudia},
  journal={arXiv preprint arXiv:2410.06062},
  year={2024}
}

@inproceedings{zahera2024cot,
  title={Generating SPARQL from Natural Language Using Chain-of-Thoughts Prompting.},
  author={Zahera, Hamada M and Ali, Manzoor and Sherif, Mohamed Ahmed and Moussallem, Diego and Ngomo, Axel-Cyrille Ngonga},
  booktitle={SEMANTICS},
  pages={353--368},
  year={2024}
}

@article{cot_paper,
  title={Chain-of-thought prompting elicits reasoning in large language models},
  author={Wei, Jason and Wang, Xuezhi and Schuurmans, Dale and Bosma, Maarten and Xia, Fei and Chi, Ed and Le, Quoc V and Zhou, Denny and others},
  journal={Advances in neural information processing systems},
  volume={35},
  pages={24824--24837},
  year={2022}
}

\appendix

\section{Appendix A}
\label{sec:appendix_B}


\begin{figure*}
    \centering
    \includegraphics[width=0.9\textwidth]{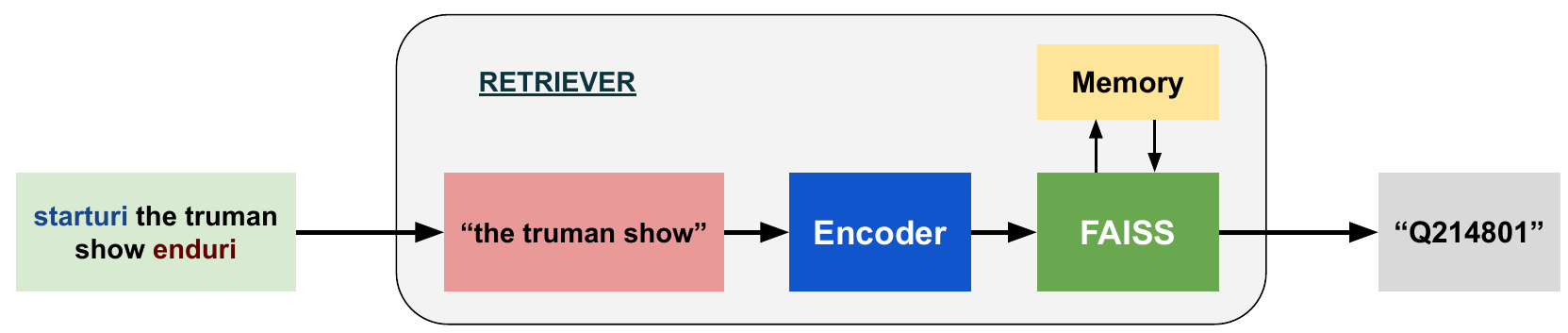}
    \caption{The retriever finds each natural language URI label in the generated intermediate query using the special tokens and retrieves the corresponding URI from memory using similarity search, as discussed in \refsec{sec:retriever}}.
    \label{fig:retriever}
\end{figure*}



\begin{figure}
    \centering
    \includegraphics[width=\linewidth]{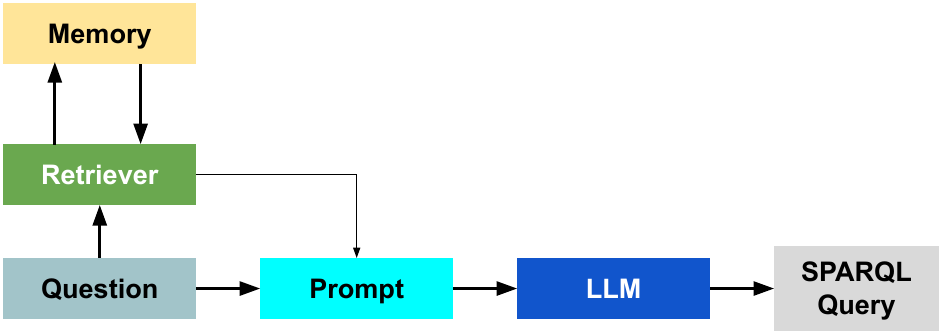}
    \caption{RAG-based SPARQL generation workflow. A retriever first fetches $k$ URIs based on latent similarity to the input question, which are then utilized by the LLM to synthesize the final query, as discussed in \refsec{sec:rag}}.
    \label{fig:rag}
\end{figure}


\begin{figure}[ht]
    \centering
    \includegraphics[width=\linewidth]{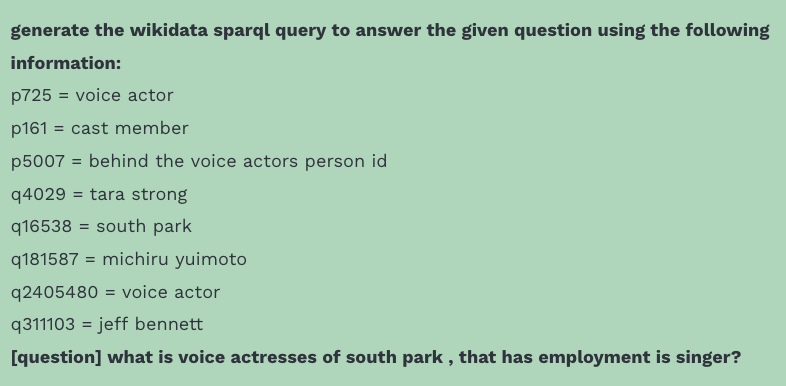} 
    \caption{The prompt for RAG includes k retrieved URIs along with the question as input to the LLM. In our study, we set $k$ to 10 in our RAG baseline.}
    \label{fig:rag_prompt}
\end{figure}

\subsection{Retrieval Augmented Generation (RAG)}
\label{sec:rag}
Retrieval Augmented Generation (RAG) \citep{knnlm, rag_og, guu2020retrieval, lewis2020retrieval} has become a popular technique for injecting external information into LLMs, helping them generate content grounded in an external knowledge base like a KG. As seen in \reffig{fig:rag}, we propose RAG for SPARQL query generation, where the question first goes to a retriever (discussed in further detail in \refsec{sec:retriever}) which returns $k$ URIs from the memory whose labels are the closest to the question text in latent space, where $k$ is a positive integer. The URIs and their labels are appended to the question to create the prompt for the LLM (see \reffig{fig:rag_prompt}). The LLM then uses this information to generate the SPARQL query corresponding to the given question.

\section{Appendix B}
\label{sec:appendix}

Unlike the evaluation approach utilized by \citet{banerjee} and \citet{qi2024enhancing}, which involves generating multiple queries for each question and selecting the first query that successfully executes and returns an answer as the predicted query, our evaluation generates only a single query from our models. 

\subsection{Hyperparameters}
\label{sec:hyperparameters}

Our T5-small model has 60,506,624 trainable parameters and was finetuned on a single 80GB Nvidia A100 GPU within 14 hours with a batch size of 128.
We used a learning rate of 0.0015 for finetuning T5 and trained for 150 epochs on LCQUAD 2.0 and 30 epochs on QALD-10. Due to the limited training data for QALD-10, we fine-tuned our versions of T5, which were pre-trained on the much larger LCQUAD 2.0 dataset. 

For our experiments with Llama 3.1 8B, we fine-tuned the Llama-3.1-8B-Instruct model, training it for 10 epochs using the AdamW optimizer with a learning rate of 0.0001, no weight decay, and a fixed seed of 42 to ensure reproducibility. Mixed precision training and quantization were enabled to improve efficiency. We employed a parameter-efficient finetuning strategy using LoRA, with a low-rank dimension of 64, a scaling factor (alpha) of 128, and a dropout rate of 0.05. The complete training process took around 3 hours and 30 minutes. Notably, the model achieved convergence at epoch 5, underscoring its rapid convergence and effectiveness in capturing key aspects of the target data.



\end{document}